\documentclass[conference]{IEEEtran}
\usepackage{tcolorbox}
\usepackage{listings}
\usepackage{xcolor}
\usepackage{multirow}
\usepackage{booktabs}
\tcbset{
	colback=gray!5,
	colframe=black!50,
	fonttitle=\bfseries,
	boxrule=0.5pt,
	arc=3pt,
	outer arc=4pt,
	coltitle=white
}
\lstdefinelanguage{bash}{
	keywords={ls, cd, mkdir, rm, cp, mv, echo, exit, run, matlab,grep},
	keywordstyle=\color{blue}\bfseries,
	basicstyle=\ttfamily\fontsize{7pt}{8pt}\selectfont,
	commentstyle=\color{gray},
	stringstyle=\color{red},
	morecomment=[l]{\#},
	morestring=[b]"
}
\lstdefinelanguage{json}{
	basicstyle=\ttfamily\fontsize{7.1pt}{8pt}\selectfont,
	showstringspaces=false,
	breaklines=true,
	frame=single,
	literate=
	*{0}{{{\color{blue}0}}}{1}
	{1}{{{\color{blue}1}}}{1}
	{2}{{{\color{blue}2}}}{1}
	{3}{{{\color{blue}3}}}{1}
	{4}{{{\color{blue}4}}}{1}
	{5}{{{\color{blue}5}}}{1}
	{6}{{{\color{blue}6}}}{1}
	{7}{{{\color{blue}7}}}{1}
	{8}{{{\color{blue}8}}}{1}
	{9}{{{\color{blue}9}}}{1}
	{:}{{{\color{teal}{:}}}}{1}
	{,}{{{\color{teal}{,}}}}{1}
	{\{}{{{\color{gray}{\{}}}}{1}
	{\}}{{{\color{gray}{\}}}}}{1}
	{[}{{{\color{gray}{[}}}}{1}
	{]}{{{\color{gray}{]}}}}{1}
	{"}{{{\color{red}{"}}}}{1}
}
\usepackage{tabularx}
\newcommand{\RomanNumeral}[1]{\uppercase\expandafter{\romannumeral#1}}

\begin{document}
%
\title{A2H-MAS: An Algorithm-to-HLS Multi-Agent System for Automated and Reliable FPGA Implementation}

\author{
	\IEEEauthorblockN{
		Jie Lei\IEEEauthorrefmark{1},
		Ruofan Jia\IEEEauthorrefmark{2},
		J. Andrew Zhang\IEEEauthorrefmark{1},
		Hao Zhang\IEEEauthorrefmark{1}
	}
	\IEEEauthorblockA{\IEEEauthorrefmark{1}
		University of Technology Sydney, Sydney, Australia\\
		Email: \{jie.lei, andrew.zhang, hao.zhang\}@uts.edu.au
	}
	\IEEEauthorblockA{\IEEEauthorrefmark{2}
		Xidian University, Xi'an, China\\
		Email: rf.jia@stu.xidian.edu.cn
	}
}
\maketitle

\begin{abstract}
Bridging the gap between algorithm development and hardware realization remains a persistent challenge, particularly in latency- and resource-constrained domains such as wireless communication. While MATLAB provides a mature environment for algorithm prototyping, translating these models into efficient FPGA implementations via High-Level Synthesis (HLS) often requires expert tuning and lengthy iterations. Recent advances in large language models (LLMs) offer new opportunities for automating this process. However, existing approaches suffer from hallucinations, forgetting, limited domain expertise, and often overlook key performance metrics. To address these limitations, we present A2H-MAS, a modular and hierarchical multi-agent system. At the system level, A2H-MAS assigns clearly defined responsibilities to specialized agents and uses standardized interfaces and execution-based validation to ensure correctness and reproducibility. At the algorithmic level, it employs dataflow-oriented modular decomposition and algorithm-hardware co-design, recognizing that the choice of algorithm often has a larger impact on hardware efficiency than pragma-level optimization. Experiments on representative wireless communication algorithms show that A2H-MAS consistently produces functionally correct, resource-efficient, and latency-optimized HLS designs, demonstrating its effectiveness and robustness for complex hardware development workflows.
\end{abstract}


%
\IEEEpeerreviewmaketitle

\section{Introduction}
Hardware design and development remain challenging and time-consuming tasks, particularly when bridging the gap between high-level software algorithms and efficient FPGA implementations. This gap becomes especially pronounced in latency- and resource-sensitive domains such as wireless communication, where stringent performance requirements demand careful architectural exploration and optimization. In practice, many of these algorithms are first prototyped and verified in MATLAB, which provides a mature ecosystem, extensive libraries, and a large body of open-source reference designs \cite{gilat2017matlab}. These MATLAB models thus serve as a natural starting point for hardware realization.
High-Level Synthesis (HLS) offers an attractive next step in this process by allowing designers to describe hardware functionality at a higher level of abstraction using C or C++ \cite{martin2009high}. Compared with traditional RTL design, HLS enables faster development and rapid design space exploration, making it an ideal intermediate representation for transforming algorithmic models into hardware implementations. However, the transition from MATLAB code to high-quality HLS code remains a significant bottleneck. Achieving efficient, resource-aware HLS designs requires substantial experience with coding patterns, pragmas, and optimization strategies, and the development cycle still involves lengthy iterations of manual tuning and verification. This challenge motivates the need for automated and reliable approaches that can directly translate MATLAB algorithms into performant HLS code, enabling efficient hardware deployment.

The recent rise of large language models (LLMs) has created new opportunities for automating hardware design and deployment. Their strong ability to generate and reason about code across multiple languages and abstraction levels makes them appealing for bridging the gap between algorithm development and hardware realization.
Recent studies have explored the use of LLMs in hardware design by fine-tuning them on Verilog-related corpora. Works such as VerilogEval \cite{liu2023verilogeval}, MG-Verilog \cite{zhang2024mg}, and VGen \cite{thakur2023benchmarking} introduced benchmarks, curated datasets, and domain-specific evaluation protocols to improve LLM performance on HDL generation tasks. While these approaches demonstrate the feasibility of adapting LLMs to hardware design, they face two key limitations. First, fine-tuning large models is computationally expensive and often impractical in real-world engineering environments. Second, recent state-of-the-art (SOTA) general-purpose LLMs (such as Claude \cite{anthropic2024claude}, ChatGPT \cite{achiam2023gpt}, and Gemini \cite{comanici2025gemini}) already exhibit strong zero-shot and few-shot capabilities, frequently outperforming smaller fine-tuned models on Verilog generation tasks \cite{liu2023verilogeval, lai2025analogcoder}. These observations suggest that domain-specific fine-tuning is not strictly necessary. A more promising direction is to unlock the latent potential of SOTA LLMs and make them stable and reliable enough to handle complex hardware design and implementation tasks.
In contrast to fine-tuning-based approaches, several recent works leverage SOTA general-purpose LLMs directly and build agent-based frameworks for hardware design. For example, VeriMind \cite{nadimi2025verimind}, HLSPilot \cite{xiong2024hlspilot}, and HDLAgent \cite{kashanaki2024hdleval} integrate testing, querying, and feedback mechanisms into an iterative code refinement loop to enhance the quality of generated HDL or HLS code. These systems represent an important step toward more autonomous hardware design. However, they still face notable limitations. First, their evaluation metrics often focus primarily on functional correctness while overlooking code quality factors such as latency, throughput, and resource utilization, which are critical for high-performance hardware. Second, because LLMs inherently suffer from hallucinations, forgetting, and limited domain expertise, a single agent is often insufficient to reliably handle the complexity of real-world, multi-stage hardware development workflows. As a result, achieving efficient and reliable end-to-end hardware implementation for complex projects remains an open research challenge.
\begin{figure}[t]
	\centering
	\includegraphics[width=\linewidth]{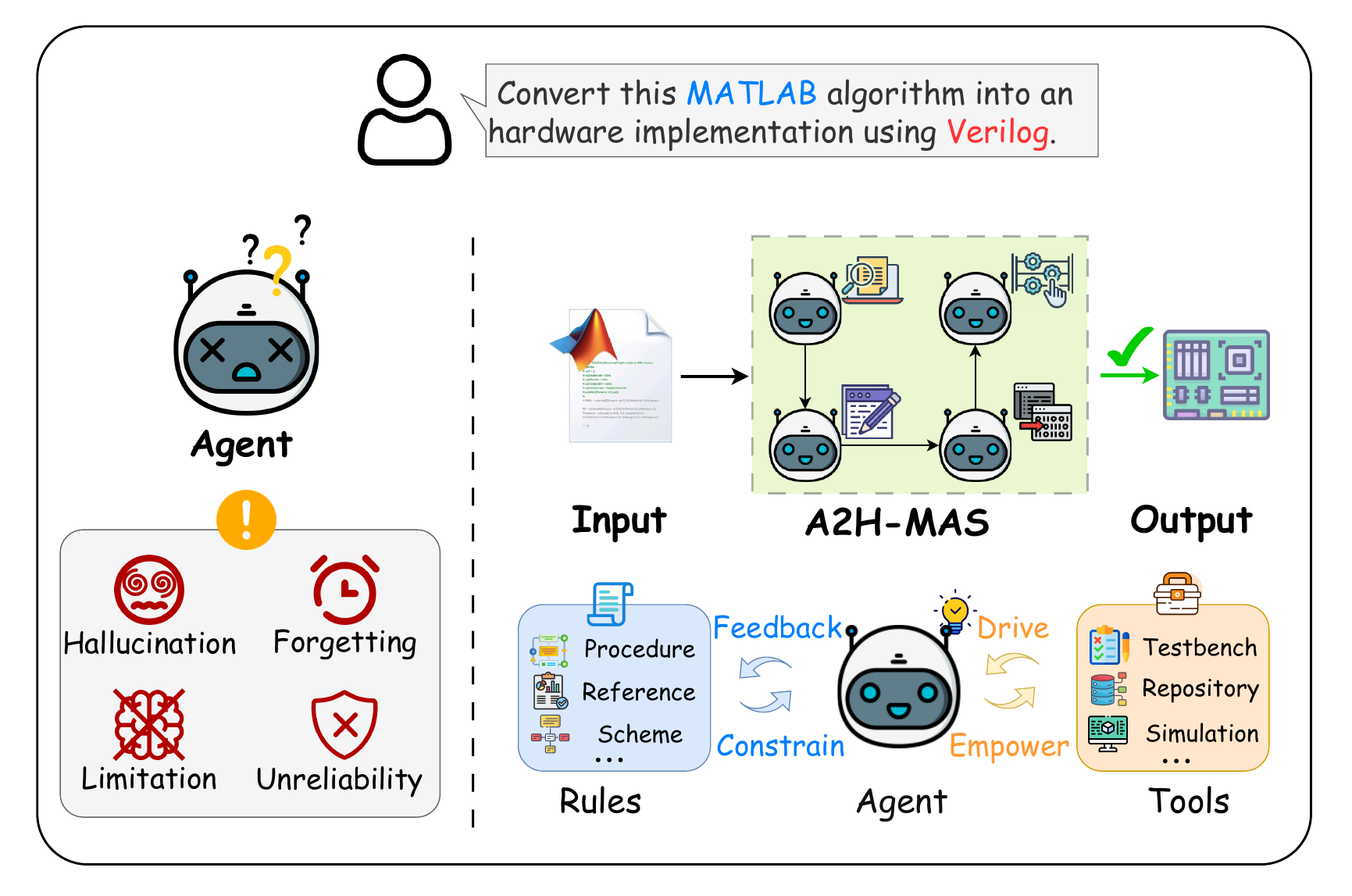}
	\caption{Compared with a single LLM agent that suffers from hallucination, forgetting, and unstable behavior, A2H-MAS decomposes the MATLAB-to-HLS-to-hardware flow into specialized agents with standardized interfaces. Agent outputs are guided by explicit rules, constrained and verified using deterministic tools, and refined through feedback, resulting in reliable and high-quality hardware implementations.}
	\label{fig1}
\end{figure}

To address these challenges, we propose \textbf{A2H-MAS} (Algorithm-to-HLS Multi-Agent System), a modular and hierarchical framework combined with a dataflow-based algorithm decomposition strategy. As illustrated in Figure \ref{fig1}, A2H-MAS transforms a MATLAB algorithm into a high-quality hardware implementation through a set of coordinated agents, each with clearly defined responsibilities. Compared with relying on a single monolithic LLM agent, which often suffers from hallucination, forgetting, and unstable behavior, A2H-MAS improves controllability and reliability by decomposing the end-to-end flow into manageable subtasks with standardized interfaces.
At the framework level, A2H-MAS adopts a multi-agent collaboration architecture in which each agent handles a specific step of the MATLAB-to-HLS-to-Verilog pipeline. The system constrains and guides agent behavior through well-defined input/output specifications, interaction rules, and execution-based validation, leveraging deterministic tools (e.g., simulators, testbench frameworks) to verify results and provide feedback. This approach reduces hallucinations, mitigates context loss, and increases reproducibility.
At the algorithmic level, A2H-MAS employs a dataflow-oriented modular decomposition to break down complex software algorithms into smaller, independent computational units. This reduces the cognitive load of individual agents, enables targeted verification and optimization, and facilitates scalable system extensions. Together, these innovations form a coherent top-down solution that systematically addresses LLM limitations and enables reliable, efficient translation of MATLAB algorithms into synthesizable HLS code and hardware implementations.

In summary, this paper makes the following contributions:
\begin{itemize}
	\item We develop A2H-MAS, a modular and hierarchical framework that performs fully automated, end-to-end conversion of MATLAB algorithms into high-quality HLS C++ code, serving as an efficient bridge toward synthesizable hardware implementations.
	\item We introduce an algorithm-hardware co-design methodology that prioritizes algorithmic transformation over pragma tuning. Appropriate algorithm selection can yield order-of-magnitude improvements in resource efficiency while preserving latency constraints. To address LLM-specific challenges such as hallucination, forgetting, and limited domain expertise, we employ a multi-agent collaboration architecture with standardized interfaces, execution-based validation using deterministic tools, and external knowledge guidance.
	\item We conduct extensive experiments on representative wireless communication algorithms, where stringent latency and resource requirements demand high-quality design solutions. The results show that A2H-MAS consistently produces functionally correct, performance-optimized, and reliable hardware implementations, demonstrating its effectiveness and stability in complex real-world engineering workflows.
\end{itemize}
\section{Related Work}
Large Language Models (LLMs) have demonstrated exceptional proficiency in code generation for high-level languages like Python and C++ \cite{anthropic2024claude, comanici2025gemini, deepseek2024deepseek, achiam2023gpt}. However, their application to Hardware Description Languages (HDLs) remains challenging due to the scarcity of high-quality training data \cite{lu2024rtllm, thakur2023benchmarking, thakur2024verigen, zhang2024mg}. While fine-tuning offers partial improvements, it often incurs significant computational costs without matching the reasoning capabilities of general-purpose models \cite{liu2023verilogeval, lai2025analogcoder}. Consequently, recent research has pivoted toward agent-based frameworks that integrate compiler and simulator feedback into the design loop. Systems such as AutoChip \cite{thakur2023autochip}, HDLAgent \cite{hdlagent}, and others \cite{chang2025data, ping2025hdlcore, novikov2025alphaevolve} utilize iterative refinement and retrieval mechanisms to enhance code quality and meet performance, power, and area (PPA) requirements.

This shift parallels the broader development of LLM-based multi-agent systems (LLM-MASs) across domains such as sociology, robotics, and software engineering \cite{park2023generative, chen2024scalable, qian2023chatdev}. Unlike single-model approaches, LLM-MASs employ structured interaction mechanisms—ranging from cooperative role allocation in frameworks like ChatDev \cite{qian2023chatdev} and MetaGPT \cite{hong2024metagpt} to adversarial debates for improving factuality \cite{du2023improving, chan2023chateval}. In the hardware domain, VeriMind \cite{nadimi2025verimind} exemplifies this paradigm by distributing verification tasks among specialized agents. By combining the generative capabilities of LLMs with the structured workflows of multi-agent collaboration, these systems offer a robust pathway for automating complex engineering tasks.

\begin{figure}[t]
	\begin{tcolorbox}[title=Prompt Template]
		
		\textbf{\large Agent Type:}
		
		Matlab\_Optimization\_Agent
		
		\textbf{\large Core Mission:}
		
		Preserve algorithm computational integrity while applying optimal HLS optimization strategy.
		
		\textbf{\large Input Parameters}
		\begin{lstlisting}[xleftmargin=0pt, framexleftmargin=0pt]
MODULE_DIR/
	|-- module_[name]_flat.m   
	|-- module_[name]_flat_tb.m       
	|-- module_definition.json   
	|-- module_[name]_in.txt
	|-- module_[name]_ref.txt
	|...\end{lstlisting}
		\textbf{\large Output Parameters}
		\begin{lstlisting}[xleftmargin=0pt, framexleftmargin=0pt]
MODULE_DIR/
	|-- module_[name]_opt.m   
	|-- module_[name]_opt_tb.m
	|...\end{lstlisting}
	\end{tcolorbox}
	\caption{Schematic of standardized input–output interfaces for agents, enabling seamless pipeline integration with minimal inter-agent coupling.}
	\label{fig:agent-prompt1}
\end{figure}
\section{Multi-Agent System Design}
Our A2H-MAS design primarily adheres to two general principles. First, the input and output interfaces of the agents are standardized to minimize inter-agent coupling. This design ensures the functional independence and integrity of each agent, thereby enabling seamless pipeline integration. Second, to enhance system stability and mitigate the impact of LLM hallucinations and memory lapses, each functionally specialized agent operates under a set of detailed guidelines. Furthermore, deterministic tools are provided to the agents for critical tasks such as verification and file operations. This setup clearly delineates the responsibilities between the LLM and the tools, thereby enabling execution-based verification to ensure the correctness and executability of the code.
\subsection{Standardized Interfaces and Seamless Integration}
Each agent is designed as an independent, self-contained module with precisely defined input and output specifications. As shown in Figure \ref{fig:agent-prompt1} , an agent receives a structured directory as input, such as the source MATLAB file, testbench, module definition, and reference data, and produces a corresponding directory with optimized or transformed artifacts. This convention allows the output of one agent to directly serve as the input to the next, forming a clear and predictable pipeline. Such design significantly reduces coupling between agents: changes to the internal implementation of an agent do not affect others, provided that the input–output contract is respected. This independence simplifies debugging and maintenance, and allows new agents to be added or existing ones replaced without reconfiguring the entire system.

\begin{figure}
	\begin{tcolorbox}[title=Prompt Template]
		\textbf{\large Phase 1: Realistic Algorithm Classification}
		
		Analyzes the module content (via grep) to classify it into one of four categories:
		\texttt{FULLY\_STREAMABLE}, \texttt{WINDOW\_STREAMABLE}, \texttt{ADAPTIVE\_STREAMABLE}, or \texttt{BLOCK\_ONLY}.  
		Each category has distinct expectations and accuracy thresholds.
		\begin{lstlisting}[language=bash]
grep -E "\bfir(filt|1|2)\b" module_[name].m
grep -E "\b(fft|ifft|fft2|ifft2)\b" module_[name].m
grep -oE "\b[a-zA-Z_][a-zA-Z0-9_]*\b" module_[name].m
| sort | uniq -c | sort -nr | head
...\end{lstlisting}
		\textbf{\large Phase 2: Apply Appropriate Strategy}
		\begin{itemize}
			\item \textbf{FULLY\_STREAMABLE:} \\
			Concrete Streaming Implementations Library. \\
			Apply proven sample-by-sample implementations using tested code from the reference library.
			\item \textbf{WINDOW\_STREAMABLE:} \\
			Sliding window streaming implementations. \\
			Use incremental update algorithms with circular buffers and accept accuracy trade-offs due to windowing effects.
			\item \textbf{ADAPTIVE\_STREAMABLE:} \\
			Fixed-parameter approximation implementations. \\
			Replace adaptive parameters with reasonable fixed values and accept approximation effects.
			\item \textbf{BLOCK\_ONLY:} \\
			Block optimization implementations.\\
			Apply explicit loop transformations and built-in elimination. Do \textbf{not} attempt streaming conversion.
		\end{itemize}
		{\huge $\cdots$}
		
		\textbf{\large Phase X: Final Validation}
		
		Use MATLAB batch execution to verify and generate test documents.
		\begin{lstlisting}[language=bash]
matlab -batch "run('module_[name]_opt_tb.m'); exit;"\end{lstlisting}
	\end{tcolorbox}
	\caption{Example of a rule-guided and tool-driven agent. Each agent follows a predefined workflow pattern and leverages deterministic tools for execution and validation, ensuring reliability and stability of the outputs.}
	\label{fig:agent-prompt2}
\end{figure}
\subsection{Deterministic Agent Operation}
To enhance the reliability of our agent and mitigate issues such as hallucinations and forgetting, we adopted a design principle that emphasizes explicit workflows and deterministic tooling for critical operations. Using the Matlab-Optimization-Agent as an illustrative example, Figure \ref{fig:agent-prompt2} highlights a subset of its workflow, presenting representative phases (classification, strategy application, and validation), each grounded in rule-based analysis and reproducible computation. Although additional phases are involved in the full pipeline, these examples capture the key idea of structuring the agent’s reasoning into modular and deterministic steps. For instance, algorithm classification is performed using deterministic grep-based feature extraction, ensuring consistent categorization across runs. Subsequent strategy selection follows a fixed decision schema that aligns implementation choices with algorithm characteristics, thereby reducing ambiguity in execution. Finally, results are validated through automated batch testing, providing a systematic safeguard against silent failures. This design approach illustrates how the framework prioritizes transparency, repeatability, and robustness in agent behavior, rather than relying solely on implicit reasoning or adaptive heuristics.

\begin{figure}
	\begin{tcolorbox}[title=module\_defination.json]
		\begin{lstlisting}[language=json]
{
	"module_name": "module_packet_detection",
	"module_id": 2,
	"algorithm_domain": "Communications",
	"function_signature": 
	{
		"name": "module2_packet_detection",		
		"inputs": [
		{ 
			"name": "filteredWaveform", 
			"type":"complex_vector", 
			"size": [26105, 2]
		},
		...
		],
		"outputs": [
		{
			"name": "startOffset", 
			"type": "int" 
		},
		...
		]
	},
	"framework_integration": 
	{ 
		"phase": 2, 
		"pipeline_position": "after_input_filtering", 
		"next_module": "module_coarse_cfo_estimation" 
	},
	"...": "XXX"
}
		\end{lstlisting}
	\end{tcolorbox}
	\caption{Illustration of the submodule configuration file, showing key fields and interface definitions.}
	\label{fig:json}
\end{figure}
\vspace{-4mm}

\subsection{Discussion}
Owing to the above design principles, our A2H-MAS exhibits the following distinct advantages:
\begin{itemize}
	\item \textbf{Complexity Reduction:} By dividing the end-to-end workflow into smaller, well-defined subtasks, each agent operates in a reduced reasoning space, mitigating error accumulation and simplifying debugging.
	\item \textbf{Improved Robustness:} Standardized interfaces and functional specialization limit uncontrolled reasoning, helping reduce hallucination and forgetting while ensuring consistent inter-agent communication.
	\item \textbf{High Extensibility:} The modular architecture allows agents to be independently added, replaced, or upgraded as new optimization strategies or hardware targets emerge, enabling continuous evolution of the system.
\end{itemize}

The framework organizes reusable capabilities into a modular skill library spanning workflow management, validation, analysis, decision-making, code transformation, documentation, and error recovery. Each skill encapsulates a well-defined capability that can be composed by multiple agents, enabling systematic code reuse and facilitating framework evolution as new optimization patterns are discovered.

\begin{figure*}[t]
	\centering
	\includegraphics[width=\linewidth]{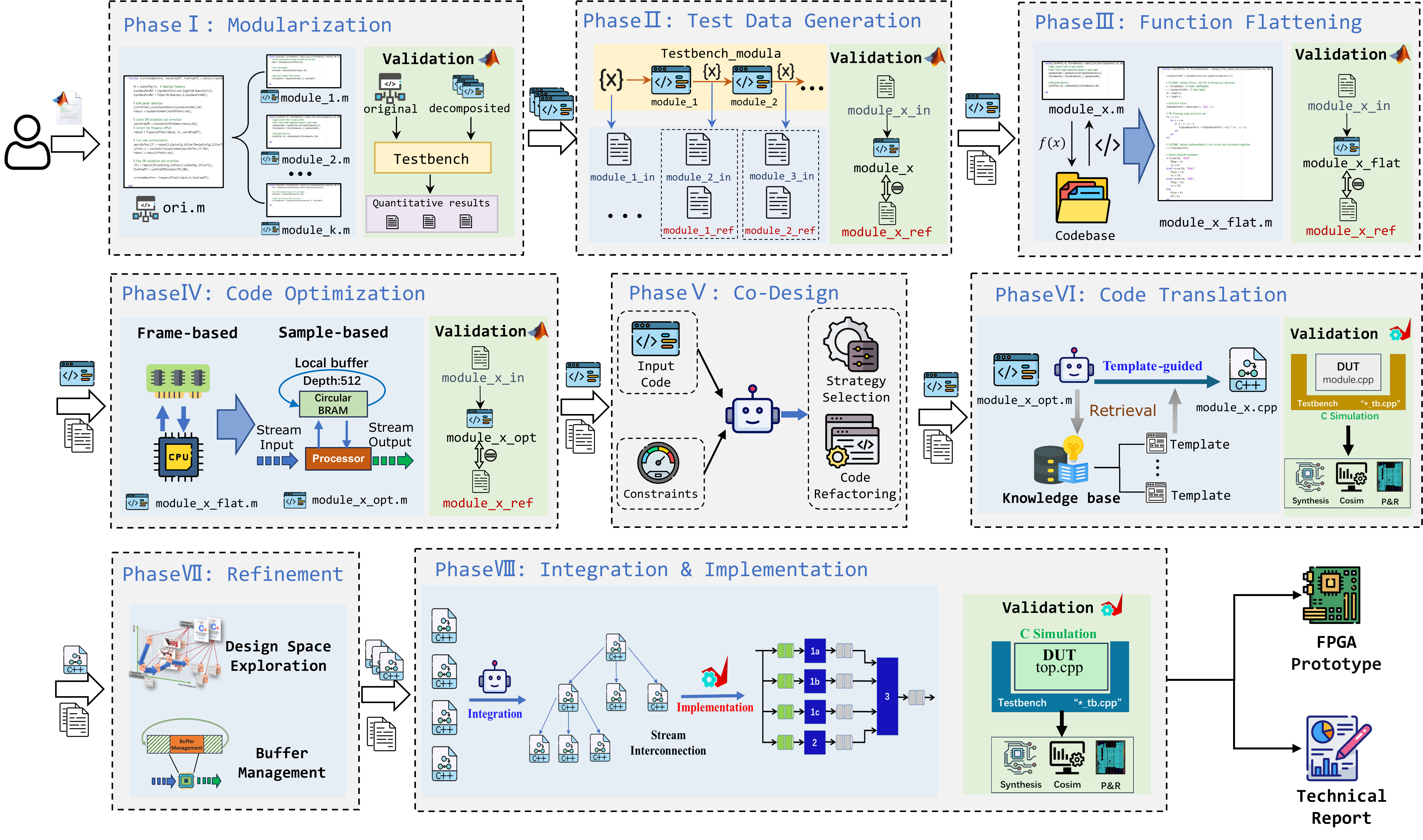}
	\caption{Overall workflow of the A2H-MAS framework. The process is organized into seven modular phases, including modularization, test data generation, function flattening, code optimization, code translation, refinement, and final integration \& implementation. Each phase is handled by a dedicated Agent responsible for a specific function, with validation mechanisms ensuring correctness, ultimately achieving a stable and efficient FPGA implementation.}
	\label{fig2}
\end{figure*}
\section{Workflow of A2H-MAS}
To systematically transform MATLAB algorithms into high-quality HLS code, A2H-MAS organizes the workflow into a sequence of modular phases, as illustrated in Figure ~\ref{fig2}. These phases cover code modularization, test data generation, optimization, translation, refinement, and final integration. Each agent follows a well-defined workflow and applies execution-based verification to ensure correctness and reproducibility.

\subsection{Phase \RomanNumeral{1}: Modularization}
The first phase of A2H-MAS is modularization, which plays a crucial role in ensuring the stability and efficiency of the entire workflow. In this phase, complex MATLAB algorithms are systematically decomposed into a set of smaller, well-defined submodules, each with a clear functional scope. This decomposition not only reduces the cognitive and computational burden on individual agents, making their outputs more stable and reliable, but also enables parallel execution of submodules wherever dependencies allow.

From a technical perspective, the modularization process strictly follows a dataflow-oriented design principle. Each submodule is defined with a minimal and standardized input–output interface, thereby avoiding the generation of redundant or overly complex intermediate results. Once modularization is complete, a structured configuration file is automatically generated for each submodule, as illustrated in Figure \ref{fig:json}. These configuration files specify key information such as the module name, interface parameters, and pipeline-related attributes, and are stored in a machine-readable format. This structured metadata provides traceability and facilitates downstream tasks such as optimization, code translation, and validation in a reproducible manner.

\subsection{Phase \RomanNumeral{2}: Test Data Generation}
The second phase of A2H-MAS focuses on test data generation, which is essential for ensuring the reliability of subsequent processing stages. For each submodule produced during Phase I, representative input–output data pairs must be available to enable functional validation, regression testing, and reproducibility.

Given that the submodules are defined according to a dataflow-oriented decomposition, their interfaces are tightly coupled: the output of one submodule directly serves as the input to the next. Moreover, these values correspond to intermediate variables of the original algorithm. Consequently, generating test data is both straightforward and systematic. We execute the original algorithm within its testbench, replace the corresponding segments with the newly defined submodules, and record all intermediate variables as they are produced. In this way, we automatically obtain both the test inputs and the reference outputs for each submodule.

All test data and reference values are stored as files using a standardized naming convention derived from the submodule configuration files (e.g., \textbf{module\_[name]\_in.txt} for inputs and \textbf{module\_[name]\_ref.txt} for reference outputs). This strict naming and storage scheme ensures traceability and guarantees that the same dataset is consistently used across all downstream phases, preserving functional correctness and stability. Final verification is performed by feeding the generated test data back into each submodule implementation and comparing the outputs against the reference values in a straightforward pass/fail manner.
\subsection{Phase \RomanNumeral{3}: Function Flattening}
The third phase of A2H-MAS is function flattening, which aims to expand MATLAB built-in or toolbox function calls into their explicit implementations within each submodule. In practical engineering projects, developers frequently rely on pre-packaged MATLAB functions to simplify algorithm development. However, these encapsulated functions present two key challenges: (i) they often require modification or optimization for efficient hardware implementation, and (ii) their abstraction can hinder LLMs from fully understanding the computational semantics of the code.

To address these challenges, we systematically traverse each submodule and identify all external function calls. Using MATLAB’s built-in \texttt{which} function, we locate the source file corresponding to each function and extract its concrete implementation. The extracted code is then inlined into the submodule, replacing the original function call. As a result, we obtain a fully flattened MATLAB file for each submodule, named according to a standardized convention (e.g., \textbf{module\_[name]\_flat.m}). This flattening process exposes the complete computational logic of the module, enabling both deterministic optimization in the following phase and improved interpretability for downstream code.

\begin{figure}[t]
	\begin{tcolorbox}[title={Example: Knowledge Library}]
		\textbf{\large Shift Register}
		
		The Shift Register pattern is commonly used for FIR filters and other parallel sample access operations. It models the hardware behavior of a delay line by locally storing the most recent $N$ samples and enabling parallel computation.
		
		\begin{lstlisting}[language=Matlab]
function y = fir_filter_opt(x, h)
 N = length(x);
 L = length(h);
 y = zeros(N, 1);
 shift_reg = zeros(L, 1); 
 % Local shift register
 for i = 1:N
  shift_reg(2:L) = shift_reg(1:L-1);
  shift_reg(1) = x(i);
  y(i) = sum(h .* shift_reg); 
  % Parallel access
 end
end
		\end{lstlisting}
		\textbf{Key benefit:} eliminating random global access and enabling fully pipelined II=1 implementation.
		
		\textbf{\large Circular Buffer}
		
		The Circular Buffer pattern is ideal for delay lines and sequential one-in-one-out memory operations, mapping efficiently to dual-port BRAM.
		\begin{lstlisting}[language=Matlab]
function y = delay_line_opt(x, delay)
 N = length(x);
 y = zeros(N, 1);
 buf = zeros(delay, 1); ptr = 1;
 for i = 1:N
  buf(ptr) = x(i);
  read_ptr = mod(ptr-delay-1, delay)+1;
  y(i) = buf(read_ptr);
  ptr = mod(ptr, delay) + 1;
 end
end
		\end{lstlisting}
		
		\textbf{Key benefit:} avoiding shifting the entire buffer and ensuring $O(1)$ update cost per sample.\\
		{\huge $\cdots$}
	\end{tcolorbox}
	\caption{Diagram of the knowledge library.}
	\label{fig:library}
\end{figure}
\vspace{-3mm}
\subsection{Phase \RomanNumeral{4}: Code Optimization}
The fourth phase, code optimization, is one of the most critical steps in determining the efficiency of the final hardware implementation. In this stage, the computation paradigm of each submodule is transformed from the frame-based and memory-centric model typical of CPU-oriented MATLAB code into a sample-based, streaming-oriented model suited for FPGA execution. While MATLAB generally stores and processes data in global memory, FPGA platforms operate on continuous data streams, requiring fine-grained, per-sample computation.

To achieve this transformation, the agent first analyzes the processing pattern of the flattened submodule, identifying memory-access behavior and data dependencies. It then consults a knowledge library containing well-established buffering and streaming patterns, including, though not limited to, registers for parallel multi-sample access, circular buffers for sequential delay lines, line buffers for two-dimensional row-wise processing, and ping-pong buffers for block-level streaming. Each pattern is associated with code templates and usage constraints, enabling the agent to apply them  through in-context learning (See Figure \ref{fig:library}). 
The knowledge library grows with each successful case, equipping the framework with \textbf{continuous learning} capabilities.
As a result, the optimized MATLAB code preserves the original functionality while restructuring the dataflow for efficient streaming, thereby improving overall performance on FPGA platforms and preparing the design for subsequent translation.

\subsection{Phase \RomanNumeral{5}: Algorithm-Hardware Co-Design}

A central observation motivating our framework is that the choice of algorithm has a larger effect on hardware efficiency than pragma-level tuning. Conventional HLS optimization focuses on loop transformations, memory partitioning, and pipelining directives. However, selecting an appropriate algorithmic approach, such as table-based implementations instead of iterative methods for transcendental functions, can reduce resource consumption by an order of magnitude.

This observation leads to our algorithm-hardware co-design principle: exploration of algorithmic alternatives should precede pragma-based optimization. The framework implements this through hierarchical strategy selection that considers algorithm-level transformations before applying HLS directives. For example, converting correlation operations to FIR-based architectures or reducing parallelism factors often yields larger resource savings than pragma tuning alone.
All algorithmic transformations are subject to latency constraints validated through RTL co-simulation, ensuring that resource reduction does not compromise throughput requirements.

\subsection{Phase \RomanNumeral{6}: Code Translation}
The fifth phase, code translation, converts the optimized MATLAB code into high-quality HLS-compatible C++ implementations. Thanks to the restructuring performed in Phase \RomanNumeral{4}, the translation process becomes relatively straightforward: the agent interprets each MATLAB operation and systematically replaces it with its functionally equivalent C/C++ construct. Particular attention is paid to ensuring that the resulting code conforms to the requirements of high-level synthesis, including the use of streaming interfaces for all modules and the preservation of input–output interface parameters as specified in the submodule configuration files.

To maximize functional correctness during early design exploration, variables are initially assigned sufficiently wide data types to avoid precision loss or overflow, with bit-width optimization deferred to later stages. Starting from this phase, the execution and verification environment transitions from MATLAB to Xilinx Vitis HLS. Each generated C++ module undergoes a complete HLS design cycle, including C simulation (csim), synthesis, and C/RTL co-simulation (cosim), to verify functional equivalence with the original MATLAB reference data produced in Phase II. This stage thus establishes a reliable bridge between algorithm-level descriptions and synthesizable hardware implementations, enabling design-space exploration and final system integration.
\begin{table*}[t]
	\small
	\centering
	\setlength{\tabcolsep}{8pt}
	\caption{Implementation Results of All Submodules for 5G NR SSB Detection and WLAN Synchronization.}
	\begin{tabular}{c c *{5}{c}}
		\toprule
		\textbf{Task} & \textbf{Implementation} & \textbf{LUTs} & \textbf{FFs} & \textbf{DSP} & \textbf{BRAMs} & \textbf{Clock (MHz)} \\
		\midrule
		\multirow{6}{*}{5G NR (SSB Detection)}
		& pssCorrelator      & 6,329 & 21,088 & 276 & 0 & 254.00 \\
		& calcThreshold      & 173   & 274    & 3   & 1 & 322.27 \\
		& peakFinder         & 1,061 & 1,439  & 0   & 0 & 279.02 \\
		& collectLocations   & 85    & 211    & 0   & 0 & 332.78 \\
		& extractSSBsig      & 155   & 148    & 0   & 4 & 269.11 \\
		\cline{2-7}
		\\[-1.7ex]
		& Top level & 8,669 & 24,216 & 279 & 7 & 292.23 \\
		\midrule
		\multirow{6}{*}{WLAN (Synchronization)}
		& filter\_detect     & 908   & 4,190  & 111 & 0  & 407.83 \\
		& coarse\_cfo        & 3,699 & 2,393  & 15  & 16 & 359.19 \\
		& fine\_timing       & 1,784 & 5,016  & 68  & 0  & 359.71 \\
		& fine\_cfo          & 3,942 & 2,755  & 16  & 29 & 358.81 \\
		\cline{2-7}
		\\[-1.7ex]
		&Top level & 10,981& 14,787 & 210 & 76 & 337.61 \\
		\bottomrule
	\end{tabular}
	\label{tab:impl_results}
\end{table*}
\begin{table*}[t]
	\small
	\centering
	\setlength{\tabcolsep}{8pt}
	\caption{Ablation Results for Modules under Various Optimization Stages.}
	\begin{tabular}{cccccccc}
		\toprule
		\textbf{Module} & \textbf{Method} & \textbf{LUT} & \textbf{FF} & \textbf{DSP} & \textbf{BRAM} & \textbf{Clock(MHz)} & \textbf{Latency} \\
		\midrule
		\multirow{3}{*}{calcThreshold} 
		& Direct     & 36,500 & 80,434 & 38 & 16 & \textit{Failed} & 6,385 \\
		& Adaptation & 685    & 1,176  & 24 & 4  & 277.09 & 6,301 \\
		& Refinement & 173    & 274    & 3  & 1  & 322.27 & 6,013 \\
		\midrule
		\multirow{3}{*}{extractSSBsig} 
		& Direct     & 4,468  & 7,071  & 0  & 24 & 265.11 & 24,890 \\
		& Adaptation & 275    & 353    & 0  & 4  & 253.29 & 12,441 \\
		& Refinement & 155    & 148    & 0  & 4  & 269.11 & 6,730 \\
		\bottomrule
	\end{tabular}
	\label{ablation}
\end{table*}
\subsection{Phase \RomanNumeral{7}: Refinement}
The sixth phase, refinement, focuses on systematically improving the performance of the HLS-generated design without altering its functional behavior or algorithmic flow. At the core of this phase is design space exploration (DSE), which enumerates candidate design points using automated scripting tools. This scripted approach allows the agent to systematically sweep through the parameter space (such as loop unrolling and pipelining factors, data type bit-widths, and array partitioning strategies) and quantitatively evaluate their impact on latency, throughput, and resource utilization.
Our DSE does not modify the structural organization of the code, but rather performs parameter-level tuning under the constraint of functional equivalence. 
In addition to parameter exploration, targeted code-level optimizations are performed, such as eliminating unnecessary variable initialization commonly found in MATLAB code, further reducing latency and improving synthesis efficiency. The overall refinement process is iterative: explore, synthesize, analyze, and adjust, until satisfactory quality-of-results (QoR) metrics are achieved.

Our refinement process mandates RTL co-simulation for latency validation rather than relying solely on synthesis estimates. This requirement ensures that resource optimizations do not inadvertently degrade throughput, as synthesis-time latency estimates can differ from actual execution cycles. This phase thus delivers a functionally consistent yet performance-optimized design, ready for final system integration.

\subsection{Phase \RomanNumeral{8}: Integration \& Implementation}
The final phase, integration and implementation, focuses on assembling all refined submodules into a unified, synthesizable hardware system. At this stage, the design transitions from module-level optimization to system-level integration, adopting a stream-based dataflow architecture that enables multiple functional units to operate concurrently. This architectural choice maximizes throughput, improves resource utilization, and ensures that the resulting hardware design can meet the demands of complex, high-performance FPGA deployments.

To orchestrate the complete system, the framework automatically generates a top-level design file (e.g., \texttt{top.cpp}) that coordinates data movement and control signaling among all submodules. The top module guarantees accurate producer–consumer communication by replicating streaming channels when needed to maintain one-to-one dataflow semantics.
Standard streaming protocols such as AXI-Stream are employed to provide low-latency communication with minimal buffering overhead. Once integration is complete, the system undergoes a full HLS verification cycle, including C simulation, synthesis, and RTL co-simulation, followed by FPGA bitstream generation and on-board validation.

\section{Experiments}
To evaluate the practicality and robustness of the A2H-MAS framework, we conducted experiments on two representative wireless communication systems: 5G NR and WLAN.
For 5G NR, an SSB detection system compliant with 3GPP standards was implemented on an NI USRP X310 platform using MATLAB, Vitis HLS, and RFNoC. To reduce DSP usage during PSS correlation, the throughput was configured to one sample every eight cycles while still meeting the 7.68 Msps requirement, demonstrating an effective trade-off between resource utilization and performance.
For WLAN, a hardware-oriented time and frequency synchronization model was developed, supporting multiple bandwidths (20/40/80 MHz) and PHY formats (Non-HT, HT, VHT, and HE). The current implementation focuses on the synchronization stage, with further baseband processing planned in future work.
In both cases, Claude Code was employed for code generation and architectural optimization, significantly accelerating development and highlighting the adaptability of A2H-MAS to diverse communication standards and hardware-efficient FPGA designs.

\subsection{Module Results}
The results in Table~\ref{tab:impl_results} reflect both the quality of the automatically generated submodules and the effectiveness of their system-level integration. For each task, A2H-MAS first decomposes the target application into functionally coherent submodules and synthesizes each with hardware-aware optimization, yielding high operating frequencies and efficient resource utilization.

More importantly, the top-level results demonstrate that A2H-MAS can automatically integrate these submodules into a complete, task-level hardware system that satisfies practical deployment requirements. For the 5G NR SSB detection task, the generated submodules are composed into a closed processing pipeline with control logic and buffering, resulting in a top-level design operating at 292.23~MHz with moderate integration overhead, where DSP usage remains dominated by the correlation-intensive \textit{pssCorrelator}. Similarly, the WLAN synchronization task achieves a post-route frequency of 337.61~MHz, with increased BRAM usage primarily due to buffering across multiple synchronization stages.

Overall, these results confirm that A2H-MAS not only produces hardware-efficient submodules through automated algorithm decomposition, but also reliably integrates them into closed-loop, task-complete systems, enabling end-to-end FPGA implementations for complex wireless communication workloads.

\subsection{Ablation Study}
To evaluate the effectiveness of the proposed framework, we conduct ablation experiments on two representative submodules, \textbf{calcThreshold} and \textbf{extractSSBsig}, analyzing their hardware quality in terms of resource utilization, timing, and latency. Three implementation strategies are compared: \textbf{Direct}, \textbf{Adaption}, and \textbf{Refinement}.
The \textbf{Direct} strategy uses a LLM to directly translate MATLAB code into HLS C++, serving as a naive baseline. \textbf{Adaption} introduces hardware-oriented restructuring at the MATLAB level prior to HLS conversion, while \textbf{Refinement} applies fine-grained optimizations to functionally correct HLS designs.

The ablation results on the \textbf{calcThreshold} and \textbf{extractSSBsig} modules consistently demonstrate the effectiveness of the proposed Adaption and Refinement stages. In the Direct configuration, MATLAB-style imperative implementations are directly mapped to hardware, leading to deeply pipelined or monolithic architectures with excessive resource usage, routing congestion, and, in the case of \textbf{calcThreshold}, failure to achieve post-route timing closure. By contrast, the Adaption stage introduces hardware-oriented restructuring, such as circular buffers, running-sum accumulators, and FSM-controlled streaming pipelines, enabling successful implementation while drastically reducing logic and memory usage. Specifically, LUT consumption is reduced from 36,500 to 685 for \textbf{calcThreshold} and from 4,468 to 275 for \textbf{extractSSBsig}, with significant latency reductions in the latter case. The Refinement stage further applies fine-grained optimizations, including fixed-point width tuning, conditional handling of uninitialized memory, and control logic simplification, resulting in minimal hardware footprints (173 and 155 LUTs, respectively), substantial latency reductions (up to 73\%), and improved post-route clock frequencies of 322~MHz and 269~MHz. These results, summarized in Table~\ref{ablation}, confirm that A2H-MAS effectively transforms high-level MATLAB algorithms into hardware-efficient and timing-robust FPGA implementations.

\section{Conclusion}
In this paper, we introduced A2H-MAS (Algorithm-to-HLS Multi-Agent System), a modular and hierarchical framework for automating the conversion of MATLAB algorithms into high-quality HLS code and reliable hardware implementations. By decomposing the end-to-end workflow into specialized agents with standardized interfaces, guided by explicit rules and deterministic tools, A2H-MAS effectively mitigates hallucination, forgetting, and instability that typically hinder LLM-based approaches. The integration of dataflow-oriented algorithm decomposition and external knowledge guidance further enhances system scalability and domain adaptability. Experimental validation on representative wireless communication algorithms demonstrated that A2H-MAS achieves functionally correct and performance-compliant hardware designs, confirming its effectiveness in practical engineering scenarios.

Looking forward, we envision extending A2H-MAS to support a broader range of algorithmic domains and hardware targets, including emerging application areas such as computer vision and signal processing. Moreover, integrating richer feedback mechanisms, adaptive optimization strategies, and larger-scale benchmark evaluations will further strengthen the robustness and generality of the framework.
\bibliographystyle{IEEEtran}
\bibliography{ref}

\end{document}